# MSRFormer: Road Network Representation Learning using Multi-scale Feature Fusion of Heterogeneous Spatial Interactions

Jian Yang[a*], Jiahui Wu[b], Li Fang[c], Hongchao Fan[d], Bianying Zhang[e], Huijie Zhao[f], Guangyi Yang[f], Rui Xin[g], Xiong You[a]

a. School of Geospatial Information, Information Engineering University, Zhengzhou 450052, China

b. College of Computer and Cyber Security, Fujian Normal University, Fuzhou 350002, China

c. Quanzhou Institute of Equipment Manufacturing, Haixi Institute, Chinese Academy of Sciences, Quanzhou 362216, China

d. Department of Civil and Environmental Engineering, Norwegian University of Science and Technology, Trondheim 7034, Norway

e. China Centre for Resources Satellite Data and Application, Beijing 100094, China

f. Henan Twenty First Century Aerospace Technology Co, Ltd., Zhengzhou 450000, China

g. College of Geodesy and Geomatics, Shandong University of Science and Technology, Qingdao 266590, China

*Corresponding author: jyangtum@qq.com or jian.yang@tum.de

**Acknowledgments**

The authors would like to thank Mr. Stefan Schestakov for his helpful discussion on the implementation of TrajRNE and sharing the data and codes.

**Funding**

This work was funded by National Natural Science Foundation of China under Grant No.42130112, No.42371479, No.41901335; and China's National Key R&D Program No.2017YFB0503500.

# MSRFormer: Road Network Representation Learning using Multi-scale Feature Fusion of Heterogeneous Spatial Interactions

**Abstract**: Transforming road network data into vector representations using deep learning has proven effective for road network analysis. However, urban road networks' heterogeneous and hierarchical nature poses challenges for accurate representation learning. Graph neural networks, which aggregate features from neighboring nodes, often struggle due to their homogeneity assumption and focus on a single structural scale. To address these issues, this paper presents MSRFormer, a novel road network representation learning framework that integrates multi-scale spatial interactions by addressing their flow heterogeneity and long-distance dependencies. It uses spatial flow convolution to extract small-scale features from large trajectory datasets, and identifies scale-dependent spatial interaction regions to capture the spatial structure of road networks and flow heterogeneity. By employing a graph transformer, MSRFormer effectively captures complex spatial dependencies across multiple scales. The spatial interaction features are fused using residual connections, which are fed to a contrastive learning algorithm to derive the final road network representation. Validation on two real-world datasets demonstrates that MSRFormer outperforms baseline methods in two road network analysis tasks. The performance gains of MSRFormer suggest the traffic-related task benefits more from incorporating trajectory data, also resulting in greater improvements in complex road network structures with up to 16% improvements compared to the most competitive baseline method. This research provides a practical framework for developing task-agnostic road network representation models and highlights distinct association patterns of the interplay between scale effects and flow heterogeneity of spatial interactions.

# 1 Introduction

Road networks play a vital role in various fields, including traffic management and urban planning. The advent of artificial intelligence, particularly deep learning, has greatly enhanced data-driven road network analysis. This approach has shown remarkable success in tasks related to road networks, such as road network matching (Yang *et al.* 2024), pattern recognition (Hou *et al.* 2024), urban traffic prediction (Zheng *et al.* 2020), and road type identification (Molefe *et al.* 2023). Unlike traditional task-driven research, data-driven road network analysis emphasizes the development of effective road network representations. This shift has transformed the research paradigm. Additionally, the widespread adoption of mobile positioning technology has generated a wealth of sensor data that helps us understand urban traffic dynamics. Researchers have begun to investigate the integration of road network data and trajectory data in creating road network representations (Chen *et al.* 2021; Mao *et al.* 2022; Schestakov *et al.* 2023; Wu *et al.* 2020). This fusion allows for a more comprehensive understanding of road networks by incorporating both their topology and dynamic traffic information. The road network representations learned by these models not only enhance the accuracy of road network analysis tasks but also contribute to a deeper understanding of urban traffic flow.

A straightforward approach to road network representation learning is to apply graph representation learning that extracts features from the neighborhood structure of roads. This involves constructing a graph embedding model for the entire road network. However, this method overlooks the inherent differences between road networks, which are geospatial data, and conventional graph data. Firstly, Graph Neural Networks (GNNs) operate under the assumption of homogeneity, meaning they consider that topologically connected nodes in a graph possess similar features and behaviors (Zhu *et al.* 2021). However, in a road network, nodes such as intersections and road segments exhibit significant spatial heterogeneity in their functions and characteristics. Thus, using traditional GNN principles may not yield an accurate or effective representation of the road network (Xiao et al. 2023). Secondly, road networks often display unique spatial structures and agglomeration patterns (Hu et al. 2023; Wu et al. 2020). These characteristics lead to traffic flow clustering in specific areas, each with its distinct features and functions (Shu et al. 2021). This spatial structure is crucial for effective road network representation learning. Unfortunately, existing research has not fully leveraged these spatial characteristics and interactions in their modeling efforts. As a result, this lack of consideration hampers the model's ability to capture dynamic traffic patterns within the road network, ultimately limiting its performance in downstream road network analysis tasks.

Recent research on road network representation learning has made significant strides in feature extraction of road network structures and the fusion of trajectory data. Some models (Wang et al. 2019; Wang et al. 2020) focus on learning representations for road intersections. These models aim to capture the surrounding road network's structure of intersections and incorporate additional information, like intersection similarity, which includes factors such as the presence of similar traffic signage. However, these approaches often rely on the assumption of homogeneity and struggle to effectively address the inherent heterogeneity within road networks. Moreover, there have been efforts to merge trajectory data representations with road networks to enhance the traffic information portrayed by graph neural networks. For instance, the model Toast (Chen et al. 2021) combines traffic context with trajectory augmentation through a Transformer module. This approach is designed to extract driving semantics from trajectories, thereby improving the overall representation of the road network in the model. Another noteworthy contribution, Jointly (Mao et al. 2022), is the joint utilization of trajectory data to create a transfer view, which fuses structural features from road network maps with contextual information from the trajectory transfer matrix. This results in an enhanced multiview map structure that guides the learning process for road network representation. Additionally, TrajRNE

(Schestakov et al. 2023) introduces a kth-order road transfer matrix using vehicle trajectory data. This method combines the transfer matrix with road network structural information to aggregate neighborhood features, ultimately computing the road network representation. Despite the interaction feature modeling these methods employ using trajectory data, they largely depend on spatial interaction features at a fixed scale (i.e., a fixed-order transfer matrix). This oversight ignores the intricate interactions among various spatial structures within the road network and their fluctuations across different scales. Consequently, this limitation hampers the models' capacity to adapt to the complex spatial structures present in road networks (Zhang et al. 2023).

To enhance road network representation and improve performance on downstream tasks, this paper explores the use of trajectory data for detecting spatial structures and extracting spatial interaction features. We treat traffic flow, as represented by trajectory data, as a type of structural signal. Our approach begins by extracting spatial interaction features from the trajectory data and incorporating them into road network modeling. This aims to address the limitations posed by the homogeneity assumption in traditional graph models. Additionally, we implement multi-scale spatial interaction feature modeling to uncover interaction patterns between different spatial regions. The primary contributions of this paper include:

(1) We propose a Graph Transformer-based road network representation learning framework called MSRFormer, which incorporates feature fusion of multi-scale spatial interactions. By analyzing structural and traffic patterns at multiple scales, the model effectively addresses flow heterogeneity and captures long-distance dependencies in road networks, thereby enhancing the overall representation of these networks.

(2) The model's effectiveness is validated using real-world datasets for two road network analysis tasks. Results show that MSRFormer outperforms baseline methods, particularly in traffic-related tasks, achieving up to 16% improvement in complex road networks compared to the top competitor.

(3) We examine how multi-scale spatial interaction features influence model learning for road network representation. Through analyses in various cities and validation of feature fusion strategies, we identify unique patterns in the relationship between scale effects and flow heterogeneity in spatial interactions.

The remainder of the paper is structured as follows: Section 2 provides an overview of the related work. Sections 3 and 4 detail the relevant definitions and present the proposed representation learning framework for road networks. Section 5 discusses the results of the comparative experiments and includes analyses. Section 6 further evaluates the effectiveness of the model design. Finally, Section 7 concludes with a summary.

## 2 Related Works

In road network representation learning, traditional methods primarily focus on the topological structure of graphs. The RFN (Jepsen *et al*. 2019) model captures the topology of a road network by creating both a node-relationship view and an edge-relationship view. It uses a message-passing mechanism to integrate these two views. However, these approaches are mainly designed for static graphs and do not take geospatial features and dynamics into account. To address this limitation, IRN2Vec (Wang *et al*. 2019) introduces geographic information and utilizes a multi-task learning framework to improve the model's spatial understanding of the road network. It employs a skip-gram model to predict contextual information about road segments, including geographic locations and intersection labels. As the complexity of road network applications increases, researchers have begun to incorporate trajectory data into the modeling process. HRNR (Wu *et al.* 2020) proposed a hierarchical graph neural network framework that models road networks at different semantic levels, covering functional regions, structural regions, and specific road segments. Despite its effective hierarchical design, HRNR does not fully leverage the significance of trajectory data in representing road networks. In terms of trajectory data utilization,

Toast (Chen *et al*. 2021) further enhances driving semantics by combining auxiliary traffic context information with a trajectory-enhanced Transformer module, thereby improving the model's representational capability. In contrast, Jiontly (Mao *et al*. 2022) constructs a transfer view using trajectory data and learns road network representations by guiding the search for orthogonal pairs based on the augmentation of contextual graphs for structural and transitional views. TrajRNE (Schestakov *et al*. 2023), on the other hand, creates a K-order transfer matrix using vehicle trajectory information. It learns road network representations by employing this transfer matrix alongside structural information to aggregate neighborhood features. Nevertheless, while these methods successfully model interaction features through trajectory data, effectively addressing the multi-scale interaction of road networks remains challenging, particularly in modeling dynamic trajectory features and hierarchical spatial structures.

In recent years, graph representation learning has emerged as a vibrant research area with a wide range of applications. Early work in this field primarily focused on shallow methods, such as DeepWalk (Perozzi *et al*. 2014) and Node2Vec (Qiu *et al*. 2018), which generate sequences of nodes through random walks and utilize skip-gram models for learning. These methods rely on a straightforward matrix decomposition framework, aiming to preserve the similarity information between nodes in the graph. In contrast, LINE (Tang *et al*. 2015) improves information retention by explicitly modeling first-order and second-order approximations. With the rapid advancements in deep learning technology, the introduction of graph neural networks (GNNs) has created new opportunities for graph representation learning. Classical models such as graph convolutional networks (GCN) (Kipf *et al*. 2016) and graph attention networks (GAT) (Veličković *et al*. 2017) have significantly enhanced the ability to model graph structures through message-passing mechanisms and neighborhood aggregation techniques. These GNN models have proven effective in node representation learning and have achieved impressive results across various tasks. However, it is important to note that many existing methods are primarily designed for general topological graphs and do not adequately consider the unique characteristics of road networks, such as complex interactions between roads and spatial structures. This limitation restricts their application potential in critical areas like transportation and urban planning. As a result, there is an urgent need to develop specialized methods for road networks that can effectively capture these distinct data features.

Graph Transformer combines the advantages of GNNs and Transformers to efficiently process graph data. It captures long-range dependencies and global features in graphs more effectively than traditional GNNs by introducing a self-attention mechanism. GraphTrans (Wu *et al.* 2021) constructs a hybrid GNN Transformer architecture. In this framework, the GNN module first extracts local context-aware representations of nodes, and then the global self-attention mechanism of the Transformer processes these locally enhanced representations. This design allows the model to simultaneously recognize both local structure and global dependencies within the graph. This capability is particularly effective for understanding the local traffic features and long-distance spatial interactions found in road networks. Graphormer (Ying *et al.* 2021) addresses the lack of graph structure awareness in standard Transformers. It innovatively incorporates various graph structure encodings, such as centrality, shortest path distance, and edge attributes, as bias terms in the computation of self-attention scores. This approach enables the attention mechanism to inherently understand the topology, node importance, and path characteristics of the graph. Building on this research, we recognize the significance of encoding and integrating dynamic interactions reflected in trajectory data into the attention mechanism to enhance the model's ability to capture the traffic characteristics of road networks. Nodeformer (Wu *et al.* 2022) focuses on efficiently processing large-scale graph data. Its core contribution is optimizing the self-attention mechanism through an anchor-point-based attentional sampling and approximation strategy, reducing the originally $O(n^2)$ computational complexity to a near-linear level. This allows the powerful global modeling capabilities of Graph Transformers to be applied to real-world large-scale graph networks with numerous nodes.

However, road networks often exhibit unique spatial structures and dynamic aggregation patterns, leading to traffic flows that form clusters with distinct characteristics and functions in specific areas. Most existing graph Transformer approaches do not fully leverage or adapt dynamically to such regional contexts. In response, we propose a dynamic region-based graph Transformer model called MSRFormer. This model first divides dynamic regions by analyzing k-order spatial interactions, then combines the powerful feature-learning capabilities of the Graph Transformer to model multi-scale spatial interactions both within and between the defined regions. The aim is to capture the complex spatial dependencies of the road network more accurately.

## 3 Preliminary

**Definition 1. Road network.** A road network can be defined as a directed graph $G = (V, A)$, $V$ denotes the set of road segments, where each node $v_i \in V$ denotes a road segment, and $A$ is the adjacency matrix $A = \{a_{ij}\}$, where $a_{ij}=1$ if a road segment $v_j$ is connected to a road segment $v_i$ while $a_{ij} = 0$ if not connected. Each node $v_i$ is associated with an initial feature vector, which represents the attributes of the corresponding road segment.

**Definition 2. Trajectory.** A trajectory $t = \{p_1, ..., p_{|t|}\}$ is a location sequence of a moving object (e.g., vehicle, person), the locations can be represented in coordinates, i.e., the $i$-th location is denoted as $p_i = (lon_i, lat_i)$, and $|t|$ is the length of the trajectory.

**Definition 3. Spatial interaction.** Spatial interaction refers to the traffic flow formed by population movement among spatial units (e.g., roads, urban regions). In urban transportation studies, spatial interaction can be measured by the volume of traffic transfer between two roads. Given a road network $G$ and a set of trajectories $T = \{t_i\}$, the spatial interaction is represented by a road transfer matrix $P$. To capture transfer dependencies between $k$ adjacent roads, we can define a road transfer matrix $P^k$ for $k$-order spatial interactions.

**Definition 4. Road network representation learning**. Road network representation learning utilizes a learning algorithm to learn a representation $R$ of each road segment from the given road network $G$ and spatial interaction $P$. In this paper, we focus on the road network representation learning problem with k-order spatial interaction $P^k$.

## 4 Methods

### 4.1 Framework Overview

In this paper, we propose MSRFormer to learn a general road network representation from road network data and trajectory data for downstream road network analysis tasks. The model architecture is shown in Figure 1, which first constructs a series of road transfer matrices using road network and trajectory data, then computes local road embeddings using neighborhood aggregation based on spatial flow convolution. Furthermore, the local road embedding and specified scale-dependent spatial interaction matrixes are fed to a feature extraction module, which uses community detection for scale-dependent region division and a GraphTransformer for feature extraction. Three of such feature extraction modules are assembled using residual connections to fuse multi-scale spatial interaction features. The final road embeddings are obtained by implementing a contrastive learning procedure.

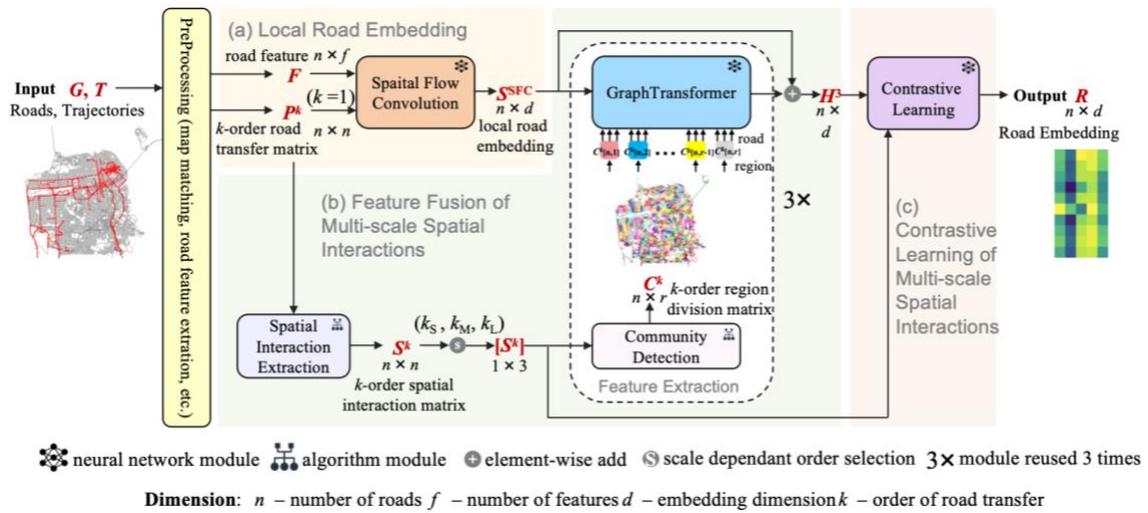

Figure 1. Overview of MSRFormer's architecture. The model learns vector representation of a road network from the road network data and trajectory data.

*4.2 Local Road Embedding*

In the initial stage of the proposed framework (see Figure1), we compute the vector representations of roads (i.e., road embeddings) via the node embedding task of a GNN, which is built on a graph model that takes road segments as nodes and road connectivity as edges. In the neighborhood aggregation of node embedding, we introduce Spatial Flow Convolution (SFC) (Schestakov *et al.* 2023) to aggregate neighbored road features based on the spatial interaction intensity between roads.

First, a road transfer matrix is constructed based on road network and trajectory data. Specifically, given a road network $G$ and a set of trajectories $T$, a map matching algorithm (Yang *et al.* 2015) was used to map a trajectory $t_i \in T$ onto the road network to obtain the corresponding road segment sequence $t_i^G = \{v_1, ..., v_j, ..., v_{|t|} \mid v_j \in V\}$. Using the adjacency matrix $A$ of the road network and the set of road segment sequences $T^G = \{t_i^G\}$ corresponding to the trajectories, the road transfer matrix for k-order spatial interactions can be computed as

$$P^k = \{p^k_{ij} \mid i, j = 1..|V|\} \tag{1}$$

and $p^k_{ij}$ is the k-order road transfer probability from road segment $v_i$ to road segment $v_j$, which is denoted as

$$p^k_{ij} = \frac{\text{count}(v_i \rightarrow v_j, T^G) + a_{ij}}{\sum_{j=1}^{|V|}(\text{count}(v_i \rightarrow v_j, T^G) + a_{ij})} \tag{2}$$

where $k$ is the number of hops between road segments, count($v_i \rightarrow v_j$, $T^G$) is the amount of traffic transfer from road segment $v_i$ to road segment $v_j$ observed in the trajectory data, $a_{ij}$ denotes the connectivity between road $v_i$ and $v_j$ in the adjacency matrix, and the denominator sums over all road $v_j$ in the road network to get the total number of transfers originating from road $v_i$ for normalization.

Then, SFC is used to perform weighted feature aggregation of the neighboring roads for each road segment based on the road transfer probability matrix $P^k$, which

helps to capture the spatial flow patterns of the local roads. The spatial flow convolution is defined as

$$S^{SFC} = \sigma\left(P^k FW\right) \tag{3}$$

where $W \in R^{f \times d}$ is a trainable weight matrix, $d$ denotes the embedding dimension, $F \in R^{n \times f}$ is the road segment feature matrix, $n$ is the count of road nodes, $f$ denotes the dimension of the initial road node features (i.e., road length, number of lanes, speed limit, road type, etc. extracted from road network data). $d$ is the preset embedding dimension of the model, which needs to be optimized and determined through cross-validation. $\sigma$ is the ReLU activation function.

Since SFC adopts a globally uniform neighborhood scale (i.e., one $k$ for all neighborhoods), it struggles to deal with the real-world traffic flow in road networks. The spatial interactions within different road network regions vary in both interaction patterns and spatial scopes, therefore one scale cannot accurately capture the differentiated spatial interactions within these regions. In addition, an oversize neighborhood (with a larger $k$) may introduce a large number of weak spatial interactions, leading to redundant and noisy aggregated information (Zhuang *et al*. 2022). This leads to the weak performance of SFC in modeling long-range spatial interactions, and it is mainly applicable to modeling local spatial interactions. Second, although SFC is capable of capturing local spatial interactions when dealing with neighborhood aggregation of complex road network structures, it falls short in handling long-distance spatial interactions, which makes it difficult to effectively capture multi-scale spatial interactions on a global scale. Therefore, we set $k = 1$ for SFC to extract local features and introduce a multi-scale spatial interaction feature fusion module to compensate for the shortcomings of SFC in capturing region-specific, long-distance dependencies and global spatial interaction features.

### *4.3 Feature Fusion of Multi-scale Spatial Interaction*

Addressing the limitations of the local road embedding, we identify the spatial interaction regions with strong spatial interactions at multiple spatial scales by analyzing the road transfer matrix, then model the global spatial interactions of the roads across multiple scales using the self-attention mechanism of Graph Transformer, and finally fuse the spatial interaction features at multiple scales with residual connections. Through the feature fusion of multi-scale spatial interactions, MSRFormer manages to model spatial interaction patterns from local to global scales while obtaining rich traffic information for the road network representation.

#### *4.3.1 Spatial Division of Multi-scale Spatial Interactions*

(1) Multi-scale Spatial Interaction Region

Spatial interaction region serves as the elemental unit of spatial interaction feature extraction, and it is often necessary to identify spatial interaction regions at multiple scales to accurately understand complex urban spatial interactions. Liu et al (2015) used taxi trajectory data to reveal the two-level hierarchical and polycentric urban structure of Shanghai city in China, demonstrating different spatial interaction patterns for short- and long-distance travel. A further study (Hu *et al*. 2023) combined the road network abstraction model and trajectory data to propose a three-level spatial structure division method to characterize a city into different functional levels, which provided a conceptual basis for the analysis of multi-scale spatial interaction features.

To facilitate the discussion, we define spatial interaction regions as regions that reveal homogeneous spatial interaction patterns. Specifically, we model such spatial interaction regions with interaction features (e.g., traffic flow intensity, interaction distance) between different spatial locations (e.g., road segment, POI) within the same region. Unlike the urban function-oriented region division, the spatial interaction region division pays more attention to traffic activity patterns. In the road network, we use a community detection algorithm to identify those road network regions with dense traffic interaction, which ensures the homogeneity of interaction patterns within the spatial interaction regions. In addition, the division scale of spatial interaction regions needs to be determined, which is used to capture the complex interaction features at different levels of road network structure. As for the number of scales, Sekihara et al. (2024) divided the road network into three levels, analyzed the functional roles of different levels of roads in travel activities with different travel distances, and introduced the hierarchical utilization ratio to reveal the relationship between the road levels and travel distances. Inspired by this study, this paper divides the spatial interactions of road networks into three scales, i.e., small-, medium-, and large-scale, according to the utilization ratios of different levels of roads and their characteristics of mobility transfer at different travel distances (as shown in Figure 2).

- **Small-scale spatial interaction**: short-distance spatial interaction within the road network, mainly reflecting the local traffic flow within the community or neighborhood. Small-scale spatial interactions usually correspond to low-order (e.g., $k$=1) road transfer features which model the mobility transfer between road segments that are directly connected.
- **Medium-scale spatial interaction**: medium-distance spatial interactions within the road network, mainly reflecting the traffic flow between urban arterials and regions. Medium-scale spatial interactions usually correspond to middle-order road transfer features which capture the traffic transfer relationship between multiple neighboring road nodes.
- **Large-scale spatial interaction**: long-distance spatial interactions within the road network, mainly involving traffic flows between urban highways and backbone roads, especially cross-regional long-distance travel. Large-scale spatial interaction usually corresponds to high-order road transfer features that model the traffic transfer between long-distance road nodes.

The definition of multi-scale spatial interaction provides a conceptual basis for the selection of multi-scale spatial interaction features, which clarifies the modeling choice of the representation of the complex spatial interaction patterns in the road network at refined resolutions.

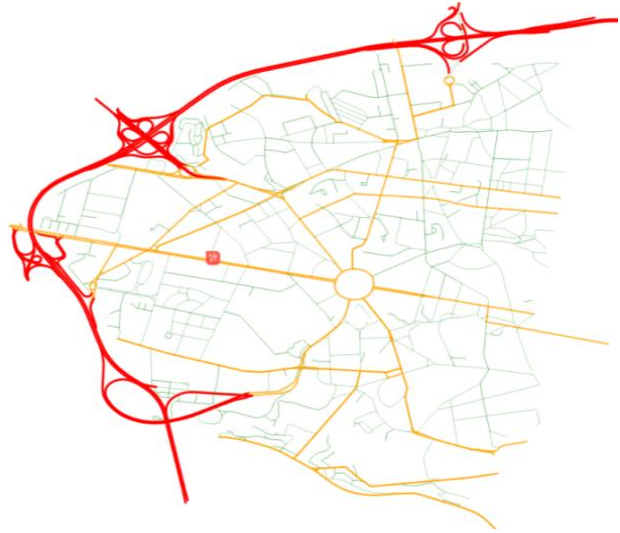

Figure 2. Hierarchy of a road network. Red-, orange-, green-lines denote highways, arterial roads, neighborhood roads respectively.

(2) K-order Region Division using Spatial Interaction Matrix

We propose a fine-grained division method for multi-scale spatial interaction regions to analyze urban road network and trajectory data at multiple scales. The method uses a community detection algorithm to divide the urban road network into multiple regions to extract the spatial interaction features in the road network more accurately. Compared with existing studies, our method reveals the spatial interactions between roads at different levels in terms of the dependency of road transfer (i.e., transfer order), and the regional extent of spatial interaction patterns. However, in the road transfer matrix $P^k$ reflecting spatial interactions, there are a large number of weak interactions between nodes, namely most of the spatial interaction values are only weakly correlated (Zhuang *et al*. 2022). Too many weak interactions will make the graph structure of the spatial interaction matrix complex and oversized while making the principled interactions between nodes cannot be effectively expressed. Therefore, we introduce modularity (Von Luxburg *et al.* 2007) to eliminate weakly connected relationships to better delineate densely connected spatial interaction regions using a community detection algorithm. Specifically, for the *k*-order road transfer matrix $\boldsymbol{P}^k = \{p_{ij}^k | i,j = 1..|V|\}$, we define the k-order spatial interaction matrix $\boldsymbol{S}^k = \{\mathbf{I}(s_{ij}^k > 0) | i,j = 1..|V|\}$, where $\mathbf{I}(\cdot)$ is the indicator function, which yields

$$p_i^k = \sum_{j \leq |v|} p_{ij}^k + p_{ji}^k \tag{4}$$

$$s_{ij}^k = p_{ij}^k + p_{ji}^k - \frac{p_i^k \times p_j^k}{2 \times \sum_{i,j} p_{ij}^k} \tag{5}$$

$$\boldsymbol{S}_{ij}^k = \begin{cases} 1, s_{ij}^k > 0 \\ 0, s_{ji}^k \leq 0 \end{cases} \tag{6}$$

Given the k-order spatial interaction matrix $\boldsymbol{S}^k$, we use the spectral clustering algorithm (Ng *et al.* 2022) to analyze the spatial interaction structure at this scale based on the interaction relationships. The spectral clustering algorithm maps the high-dimensional structure of the graph to a low-dimensional feature space through Laplace matrix feature decomposition. The reduced low-dimensional space better reflects the relative relationships and similarities between nodes, making clustering in this space more accurate and stable (Von Luxburg *et al.* 2007). The algorithm first calculates the diagonal matrix $\boldsymbol{D}^k$ of $\boldsymbol{S}^k$, which yields the Laplacian matrix $\boldsymbol{L}^k = \boldsymbol{D}^k - \boldsymbol{S}^k$. The eigenvalue decomposition of the Laplacian matrix $\boldsymbol{L}^k$ is performed by calculating the eigenvectors corresponding to the first *d* smallest eigenvalues of the Laplacian matrix $\boldsymbol{u}_1, ..., \boldsymbol{u}_d$, where $\boldsymbol{u}_i \in \boldsymbol{R}^{n \times 1}$ denotes the eigenvalue of the node in that feature dimension, and these eigenvectors are formed into a matrix $\boldsymbol{U} \in \boldsymbol{R}^{n \times d}$. The K-means algorithm is applied to the matrix $\boldsymbol{U}$ to obtain the division of the region $\boldsymbol{C}^k \in R^{n \times r}$ under the k-order spatial interaction, where

$$C^k[n,r]=\begin{cases}1, & n\in r\\0, & \text{else}\end{cases} \qquad (7)$$

where *n* denotes the road and *r* denotes the region to which it belongs. Similarly, using the K spatial interaction matrices $S^1$, ..., $S^K$, the region division under different scales of spatial interaction $\{C^1, ..., C^K\}$ can be obtained. By applying the community detection algorithm at multiple scales, the intrinsic structure in the road network can be extensively explored while providing enhanced pattern inputs for the extraction of multi-scale interaction features.

(3) Scale Selection of Spatial Interactions

After obtaining the spatial interaction region at specific scales using the spatial interaction matrix, we need to determine the order *k* of spatial interaction that aligns with the definition of spatial interaction at small, medium, and large scales. This will allow the model to better extract multi-scale spatial interaction features. Since roads at different levels of the network exhibit varying traffic flow characteristics, we aim to select an appropriate range for the interaction order. This ensures that the model effectively aggregates and captures traffic features at different scales while minimizing information redundancy and noise interference in spatial interaction. Specifically, we define the spatial interaction order range of the road network by analyzing the relationship between the spatial interaction orders of roads at different levels.

In defining the spatial interaction scales of the road network, we categorize urban roads into three levels: highways, which facilitate long-distance, cross-regional traffic; urban trunk roads, which support medium-distance inter-regional traffic; and residential roads, which serve short-distance intra-community traffic. To analyze the traffic transfer characteristics among these three road levels, we randomly sampled 1,000 trajectories from the trajectory data of two experimental areas (Porto and San Francisco). We then generated a sequence of road network paths corresponding to these trajectories through road network matching and statistically examined the distribution of interaction orders and transition patterns among roads at different levels. The experimental statistical results, depicted in Figure 3, reveal that while different cities exhibit similar road level interaction order relationships, the distribution characteristics of these orders vary between the cities. In Porto, small-scale spatial interactions are primarily concentrated in orders 1-2, medium-scale interactions in orders 3-5, and large-scale interactions in orders 6-9. In contrast, San Francisco's small-scale interactions fall within orders 1-3, medium-scale interactions within orders 4-6, and large-scale interactions within orders 7-8. Based on these findings, we determine the range of interaction orders corresponding to specific spatial interaction scales, ensuring that the model can accurately extract multi-scale spatial interaction features from different cities.

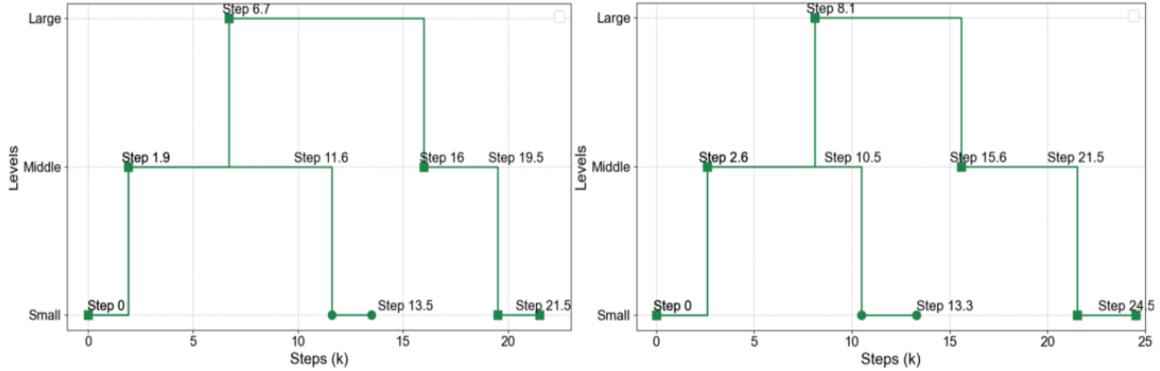

Figure 3. Orders of spatial interaction between roads with different levels in urban road network (left: Porto, right: San Francisco).

*4.3.2 Feature Extraction of Spatial Interaction using Graph Transformer*

The *k*-order region division has generated a diverse set of spatial interaction regions, denoted as $\{C^1, ..., C^K\}$, each characterized by unique distance interactions. Each region $C^k$ illustrates the traffic flow patterns and spatial structure between roads at a specific scale, offering insights into the spatial interaction characteristics of the road segments within that region. It is important to note that the dependence of spatial interaction between roads varies at different scales. For instance, roads situated in urban centers primarily facilitate short-distance travel within the community. As the scale of the road network increases, these community roads begin to connect with surrounding urban arterial roads to accommodate inter-regional traffic flows. However, graph neural networks that rely on graph topology and message passing for feature computation typically perform neighborhood aggregation locally. This approach can overlook the dependencies between distant nodes and may not adequately capture the multi-scale spatial interaction features.

To address this limitation in road embedding, we propose a method for extracting multi-scale spatial interaction features based on a Graph Transformer. Unlike message-passing-based graph neural networks, Graph Transformers utilize a self-attention mechanism to model relationships between any node pairs within the graph, even if there is no direct connection between them. This capability provides significant advantages in capturing the global dependencies of road segments in a region along with the multi-scale spatial interaction patterns. As a result, the model can effectively learn local interaction features while also uncovering larger-scale spatial interaction features within the road network.

For each analysis scale, we select the *k*-order road transfer matrix $\boldsymbol{P}^k$ to delineate the spatial interaction region $\boldsymbol{C}^k$. This allows us to segment the road network into a corresponding set of subgraph regions. We model the spatial interactions of each subgraph region using a Graph Transformer. This model captures the spatial dependencies between roadway nodes on a global scale through a self-attention mechanism, which enables the extraction of spatial interaction features at the distance scale *k*. The attention mechanism identifies global dependencies by calculating the similarity between input features. This process is centered around three matrices: query (*Q*), key (*K*), and value (*V*), which are derived from the linear transformation of the input features. Attention weights are computed using the dot product of the query and key matrices. These weights are then applied to sum the values, resulting in a weighted representation of each feature. This approach allows us to flexibly capture global information and interactions between features. Specifically, the $\boldsymbol{Q}$, $\boldsymbol{K}$, and $\boldsymbol{V}$ matrices can be obtained through the linear transformation of the input feature matrix $\boldsymbol{H}$ from the previous layer, which yields

$$Q = H^{l-1}W_Q, K = H^{l-1}W_K, V = H^{l-1}W_V \tag{8}$$

where $H^{l-1}$ represents the input feature matrix of the current layer $l$. The matrices $W_Q$, $W_K$, and $W_V \in R^{d \times d}$ are the linear projection matrices that correspond to queries, keys, and values, respectively. For the case when $l = 0$, $H^0$ refers to the output from the modular spatial flow convolution $S^{SFC} \in R^{n \times d}$. From this, we can define the attention score matrix $W_{att}$.

$$W_{att} = \text{softmax}(\frac{QK^T}{\sqrt{d}}), \quad H' = W_{att}V \tag{9}$$

where $d$ represents the vector dimension. The embedding representation of the road segment nodes is updated through a weighted summation of the values $V$ using the attention score matrix.

To enhance the model's ability to integrate spatial interaction features across different distance scales, we introduce a bias term derived from the $k$-order transfer matrix in the Graph Transformer. For pairs of road segment nodes $(v_i, v_j)$ within a subgraph of the road network, segmented by region nodes at distance scale $k$, the embeddings of nodes $v_i$ and $v_j$ are represented as $h_i$ and $h_j$, respectively. A learnable scalar bias is assigned to these nodes based on their $k$-th order transfer probabilities $p_{ij}^k$. The function $b_\varphi$ computes a specific learnable bias for an attention head using the $k$th-order transfer probability $P^k(v_i, v_j)$ of the node pair. This function enables the model to learn a different spatial dependency for each attention head. This learned bias term is summed element-by-element with the original attention score computed based on the node features, and then Softmax normalized for node $v_j$ to obtain the attention score matrix term $W_{att}[i, j]$. As a result, the attention score matrix is rewritten as follows.

$$W_{att}[i, j] = \text{softmax}_j \left( \frac{(h_i W_Q)(h_j W_K)^T}{\sqrt{d}} + b_\phi \left( P^k(v_i, v_j) \right) \right) \quad (10)$$

With the spatial interaction encoding mentioned above, the Graph Transformer can model the road nodes in the subgraph area from a global perspective and capture the spatial interaction dependency information of the road network at that distance scale.

*4.3.3 Feature Fusion of Multi-scale Spatial Interactions using Residual Connections*

To fuse the spatial interaction features across different scales, we implement residual connections between each Graph Transformer layer. Specifically, the road node features from various scales are processed through the Graph Transformer layer corresponding to their scale region. The output features of each layer are then fused using Residual Connections, allowing us to combine the output features from the spatial interaction feature extraction module across all layers. This process results in a road node embedding representation that captures multi-scale spatial interaction information. The node embedding representation in each layer is updated according to the following formula:

$$H^l = \text{GraphTransformer}(H^{l-1}) + H^{l-1} \quad (11)$$

where $H^l$ represents the node representation after it has been updated by the *l*-layer spatial interaction feature extraction module. A residual connection is established using the "+" operation, meaning that the input features $H^{l-1}$ from the previous layer are added to the output of the feature extraction module of the current layer. This approach facilitates multi-scale interaction feature fusion. By employing a feature fusion strategy based on residual connections, we achieve cross-layer fusion of multi-scale spatial interactive features. This method also allows for the transfer of information related to spatial interaction features across different scale modules. As a result, the model can more effectively integrate spatial interactive features of varying scales when navigating complex road network structures. For instance, in transportation hubs, this technique enables the learned embeddings to embody small, medium, and large spatial interaction features. The fusion of multi-scale interaction feature node representations, enhanced through multi-layer residual connections, more accurately captures the intricate spatial interactions within the road network. Consequently, this improves the overall performance of the road network representation model.

### *4.4 Contrastive Learning of Multi-scale Spatial Interactions*

To determine the parameters of the network model (such as SFC parameters and attention weights of the feature fusion module) that enable better capture of multi-scale spatial interaction features, we employ a contrastive learning algorithm for training. Contrastive learning is a self-supervised learning method that does not require labeled data; instead, it learns road node embedding representations by constructing pairs of positive and negative samples. Given that the spatial interactions within the road network vary in intensity and scale, we generate positive and negative samples of these spatial interactions based on the spatial interaction matrix $S^k$, which is derived from the road transfer matrix $P^k$ at different scales. Specifically, node pairs exhibiting spatial interactions in $S^k$ are designated as positive samples $(v_i, v_j)^{k+} \in E^+$, which are utilized to extract strong spatial interaction features at distance scale $k$. Conversely, negative samples are formed by randomly selecting node pairs from $S^k$ that do not exhibit spatial interactions, i.e., $(v_i, v_j)^{k-} \in E^-$, representing weaker or non-existent spatial interaction features at distance scale $k$. By constructing positive and negative samples across multiple scales, we ensure the model fully leverages the spatial interaction features available at different scales. Accordingly, we can design

the contrastive learning loss function for multi-scale spatial interactions as

$$L = -\frac{1}{|E^+|} \sum_{(v_i,v_j)^{k+} \in E^+} \log\left(\sigma\left((h_i^l)^\cdot, h_j^l\right)\right) - \frac{1}{|E^-|} \sum_{(v_i,v_j)^{k-} \in E^-} \log\left(1 - \sigma\left((h_i^l)^\cdot, h_j^l\right)\right) \quad (12)$$

where $E^+$ and $E^-$ denote the sets of positive and negative spatial interaction samples, respectively, $h_i^l$ and $h_j^l$ represent the embedding representations of nodes $v_i$ and $v_j$, and $\sigma$ is the Sigmoid activation function. The loss function is a binary cross-entropy loss function that implements contrastive learning to guide the self-supervised learning of the model. Specifically, its purpose is to help the model learn spatial interaction features of nodes at different scales by maximizing the similarity between positive sample pairs while minimizing the similarity between negative sample pairs. This ensures that the road node embeddings effectively capture spatial interaction features at both the local neighborhood and global scales. During the model training process, the SFC module and the multi-scale spatial interaction feature extraction module work together to optimize the loss function. The iterative optimization of the contrastive loss is achieved by iteratively updating the model's convolutional kernel parameters and attention weights. This approach enables the model to learn representations of road node embeddings that incorporate multi-scale spatial interaction features.

## 5 Experiments

### 5.1 Experiment Setting

#### 5.1.1 Dataset

To assess the effectiveness of the proposed method, we conducted a comparative analysis using trajectory datasets from Porto and San Francisco (Schestakov *et al.* 2023), along with road network data extracted from OpenStreetMap. The selected road networks for both cities differ significantly in terms of quantity and density of roads. San Francisco has a larger scale in raod quantity, while Porto is more sparse in road density, providing an practical case to evaluate the model's adaptability across different data scenarios. Data preprocessing is needed for both he road network and trajectory data for the experiment. The preprocessing involved removing trajectories that extended beyond the city boundaries, deleting those with fewer than 10 points, and matching trajectories to the road network to obtain sequences of road segments. Table 1 presents the statistical information of the experimental dataset after this preprocessing step.

Table 1. Statistics of the experiment dataset

|  | Statistics | Porto | San Francisco |
|---|---|---|---|
| Road Network | #intersection | 5376 | 9928 |
|  | #road segment | 11368 | 27428 |
|  | avg. degree | 2.4 | 3.1 |
| Trajectory | #GPS point | 74,269,739 | 1,544,234 |
|  | #trajectory | 1,544,234 | 406,456 |
|  | coverage | 97.9% | 94.4% |

*5.1.2 Evaluation Tasks*

We select two common tasks in road network analysis: road label classification and traffic inference, to evaluate the proposed method.

- **Road Label Classification (RLC)** is a task that involves identifying the type of road segment based on its characteristics. We collected road type labels (such as freeways and residential streets) from OpenStreetMap and subsequently merged and relabeled them into five categories (Molefe *et al*. 2023), as shown in Table 2. To evaluate the classification accuracy of our model, we used two metrics: Micro-F1 (Mi-F1) and Macro-F1 (Ma-F1). The Mi-F1 metric summarizes the true instances and predicted instances across all categories to calculate overall Precision and Recall. The final Ma-F1 value, on the other hand, is determined by calculating the Precision and Recall based on the aggregate of all road categories, denoted as

$$\text{Mi-F1} = (2\text{Precision}_{micro} \times \text{Recall}_{micro})/(\text{Precision}_{micro} + \text{Recall}_{micro}) \quad (13)$$

$$\text{Precision}_{micro} = \sum_{i=1}^{C} \text{TP}_i / \sum_{i=1}^{C} (\text{TP}_i + \text{FP}_i) \quad (14)$$

$$\text{Recall}_{micro} = \sum_{i=1}^{C} \text{TP}_i / \sum_{i=1}^{C} (\text{TP}_i + \text{FN}_i) \quad (15)$$

where $\text{TP}_i$, $\text{FP}_i$, and $\text{FN}_i$ represent the number of true positive, false positive, and false negative classification results for category $i$, respectively, and C is the total number of classification categories. The Macro F1 (Ma-F1) metric treats the importance of all categories equally by calculating the $\text{F1}_i$ value for each category separately and then averaging these values, which can be expressed as

$$\text{Precision}_i = \text{TP}_i/(\text{TP}_i + \text{FP}_i), \text{Precision}_i = \text{TP}_i/(\text{TP}_i + \text{FN}_i) \quad (16)$$

$$\text{Ma-F1} = \frac{1}{C}\sum_{i=1}^{C} \text{F1}_i \quad (17)$$

$$\text{F1}_i = (2\text{Precision}_i \times \text{Recall}_i)/(\text{Precision}_i + \text{Recall}_i) \quad (18)$$

Table 2. Label classification of OSM roads in the experiment

| Class | Road labels |
|---|---|
| Class1 | primary, primary_link, motorway, motorway_link, trunk, trunk_link, secondary, secondary_link |
| Class2 | road, unclassified |
| Class3 | tertiary, tertiary_link |
| Class4 | residential |
| Class5 | living street |

- **Traffic Inference (TI)** is a regression task that involves predicting the speed of traffic on each road. We calculate the average speed of each road using actual trajectory data, which serves as the true value. To evaluate the accuracy of our predictions, we utilize two metrics: the root mean square error (RMSE) and the mean absolute error (MAE) with the unit of km/h.

To accomplish the two tasks, specific task heads (module adapters tailored for each task) are required alongside the road embeddings generated by the proposed model. In the road label classification task, a classifier composed of a fully connected layer followed by a softmax layer is employed to predict road labels based on the road embeddings. For the traffic inference task, a multilayer perceptron (MLP) regression model is utilized to estimate the speed values of road segments using the road embeddings.

Furthermore, three types of baseline methods were utilized in the comparison experiments, including basic GNN methods, road network representation learning (RNRL) methods and trajectory-enhanced road network representation learning (T-RNRL) methods.

- For basic GNN methods, we selected GCN (Kipf *et al.* 2016) and GAT (Veličković *et al.* 2017) as the baselines and employed a graph reconstruction task for model training in an unsupervised manner.
- For the RNRL methods, we chose RFN (Jepsen *et al.* 2019) and IRN2Vec (Wang *et al.* 2019) as our baselines. RFN utilizes multiple views to model the relationships between nodes and edges in the road network graph. In contrast, IRN2Vec adopts a multi-task learning framework using a skip-gram model to predict geographic locations and intersection labels within the context of a specified road segment.
- For the T-RNRL methods, we selected HRNR (Wu *et al.* 2020), Toast (Chen *et al.* 2021), TrajRNE (Schestakov *et al.* 2023) and JGRM (Ma *et al.* 2024) as the baselines. HRNR proposes a GNN-based framework for hierarchical modeling at different semantic levels of a road network, including functional areas, structural areas, and road segments. Toast leverages auxiliary traffic context information to train the skip-gram model. Additionally, the trajectory data is extracted using the Transformer module for trajectory augmentation to capture driving semantics on the road network. TrajRNE employs Spatial Flow Convolution (SFC) and Structural Road Encoder (SRE) to integrate trajectory data and learn a representation of the road network. JGRM explores the motion information in GPS trajectories and learns representations of GPS data and road-based routes jointly through self-supervision.

*5.1.3 Model Implementation*

For the local road embedding, we employs one layer of spatial flow convolution and three layers of stacked spatial interaction feature extraction modules, with the $f$ set to 12 for road label classification and 26 for traffic inference in Porto, 19 and 33 respectively in San Fransisco. Each layer of the spatial interaction feature extraction module utilizes 4 attention heads, with the embedding dimension set to 208. For the spectal clustering algorithm, the spectral embedding dimension $d_s$ is set equal to the number of target clusters $r$, so that it can retain the optimal subspace structure information that is sufficient to divide $r$ regions. In the experiments, $r$ is set to 300 for the Porto dataset and 800 for the San Francisco dataset, which is larger and more complex.

The model training is conducted over 1000 epochs, and the model's learning rate is set to 0.001. During the pre-training phase, we use self-supervised contrastive learning. After training, we extract 208-dimensional road segment representation vectors from the pre-trained model. These vectors are then input into two downstream tasks: road label classification and traffic prediction, which are evaluated using five-fold cross-validation. Specifically, for each validation, the dataset is randomly divided into five equal parts. Four parts are used for training, while the remaining part serves as the test set. This process is repeated five times, with a different part designated as the test set each time, and the average of the five test results is calculated. Additionally, during the training process, we randomly select 10% of each training set to serve as the validation set, optimizing the model parameters for multi-scale order $k$ selection. Adam optimizer is used in both model pre-training and fine-tuning phases.

*5.2 Experiment Results and Analysis*

The experimental results in Table 3 indicate that the proposed MSRFormer significantly outperforms the baseline methods across both tasks on the Porto and San Francisco datasets. The enhancements demonstrate MSRFormer's effectiveness in both tasks. Furthermore, the traffic-related task benefits more from incorporating trajectory data, resulting in greater improvements in complex road network structures. The significant improvement (i.e., up to 26%) in road label classification on the Porto dataset is due to the Ma-F1 metric's sensitivity to class-wise model performance, and Porto exhibits more distinct hierarchical and irregular road network patterns compared to San Francisco.

Overall, T-RNRL methods, such as MSRFormer, TrajRNE, and HRNR, outperform the RNRL methods that only utilize road network information, such as RFN and IRN2Vec, based on experimental results. For instance, the RNRL methods RFN and IRN2Vec achieve only 0.498 and 0.487 in the Mi-F1 score for the road label classification task on the Porto dataset, as well as 15.088 and 17.366 in MAE for the traffic speed inference task, respectively. These results are significantly inferior to those of methods that incorporate dynamic traffic information. Additionally, while the basic GNN methods such as GCN and GAT display consistent performance across tasks, their overall results are still not as favorable as those of the proposed T-RNRL method. As for the most recent baseline, JGRM, the baseline model first fails to run on a mid-range GPU (i.e., NVIDIA RTX 4060 8GB) as it consumes larger GPU memory. When running MSRFormer and JGRM on a professional GPU (i.e., NVIDIA A800 80GB), MSRFormer still gets the best results in nearly all cases.

Table 3. Experiment results of MSRFormer and baselines.

| Model / Task | Porto | | | | San Francisco | | | |
|---|---|---|---|---|---|---|---|---|
| | RLC | | TI | | RLC | | TI | |
| | Mi-F1 ↑ | Ma-F1 ↑ | MAE ↓ | RMSE ↓ | Mi-F1 ↑ | Ma-F1 ↑ | MAE ↓ | RMSE ↓ |
| GCN | 0.660 | 0.411 | 14.175 | 20.361 | 0.663 | 0.070 | 10.435 | 16.294 |
| GAT | 0.651 | 0.393 | 14.238 | 20.388 | 0.676 | 0.131 | 10.113 | 15.730 |
| RFN | 0.498 | 0.087 | 15.088 | 20.908 | 0.672 | 0.125 | 10.068 | 15.660 |
| IRN2Vec | 0.487 | 0.055 | 17.366 | 23.176 | 0.658 | 0.057 | 13.336 | 19.307 |
| HRNR | 0.540 | 0.132 | 13.733 | 19.798 | 0.692 | 0.147 | 10.068 | 15.265 |
| Toast | 0.440 | 0.206 | 13.793 | 19.543 | 0.662 | 0.068 | 10.122 | 15.631 |
| TrajRNE | <u>0.682</u> | <u>0.496</u> | <u>13.228</u> | <u>19.215</u> | <u>0.759</u> | <u>0.475</u> | <u>8.437</u> | <u>13.334</u> |
| JGRM | OOM | OOM | OOM | OOM | OOM | OOM | OOM | OOM |
| MSRFormer | **0.743** | **0.626** | **12.024** | **17.624** | **0.798** | **0.505** | **7.063** | **11.070** |
| 1st outperforms 2nd | 8.94% | 26.21% | 9.10% | 8.28% | 5.14% | 6.32% | 16.29% | 16.98% |
| JGRM (+) | <u>0.731</u> | <u>0.577</u> | **11.881** | <u>18.160</u> | <u>0.760</u> | <u>0.269</u> | <u>8.521</u> | <u>13.223</u> |
| MSRFormer (+) | **0.737** | **0.622** | <u>12.230</u> | **17.611** | **0.789** | **0.510** | **7.037** | **11.141** |
| 1st outperforms 2nd | 0.82% | 7.80% | 2.85% | 3.02% | 3.82% | 89.59% | 17.42% | 15.75% |

Note: **Bold** font and <u>underline</u> indicate 1st and 2nd best results, and OOM means out of memory while using a mid-range GPU. The cross "+" indicates a model running on a professional GPU with larger memory size.

The success of the proposed method can be attributed to its effective utilization of the multi-scale spatial interaction features inherent in the road network. By combining trajectory data with road network data, this approach enhances the granularity of spatial interaction feature extraction while effectively capturing dynamic traffic patterns. This fusion strategy not only increases the model's predictive accuracy but also improves its robustness when handling unbalanced data and dynamic prediction tasks, resulting in superior performance across all metrics.

*5.3 Ablation Study*

To better understand the contribution of each module in our proposed method, we conducted ablation experiments using two tasks from the Porto dataset. This involved removing or replacing key modules in the method and evaluating their impact on the experimental results.

**Ours-w/o-SFC (A1)**: As illustrated in Figure 4, the model designated as A1 was used to assess the effectiveness of the SFC module in aggregating local neighborhood features. In this case, the SFC module was replaced with a linear layer. The results indicate that removing this module led to a decrease in the model's performance on traffic inference and road classification tasks. This suggests that the SFC module plays a crucial role in effectively aggregating local neighborhood features and providing improved local information for the multi-scale interaction feature module.

**Ours-w/o-IRP (A2)**: Similarly, the model A2 was employed to evaluate the impact of the spatial interaction region division on model performance. In this instance, the road transfer matrix-based region division strategy was replaced with a region division strategy that relies on the road network structure (similar to HRNR). The findings revealed that modifying this module also resulted in a decline in model performance for both tasks. This indicates that the spatial interaction region division is essential for the model to capture the complex interactions occurring in different areas of the road network.

**Ours-w/o-MSIF (A3)**: The model A3 eliminates the multi-scale interaction feature extraction module, retaining only the single-scale spatial flow convolution. The results indicate that removing the multi-scale spatial interaction fusion module leads to a dramatic decrease in the model's performance on both tasks, particularly in the traffic inference task, where the error significantly increases. The multi-scale interaction feature extraction module is crucial for capturing spatial interaction features at various scales. This module enhances node representation and greatly improves the accuracy of road traffic predictions by integrating the dynamic interaction region division strategy with global Graph Transformer modeling.

**Ours-w/o-RMD (A4)**: On the other hand, the model A4 removes the residual connections within the multi-scale spatial interaction module to evaluate the effectiveness of these connections for fusing multi-scale features. The absence of residual connectivity resulted in a substantial decline in model performance, especially in the road classification task. This finding confirms the importance of residual connections in enhancing model stability and facilitating deep feature fusion.

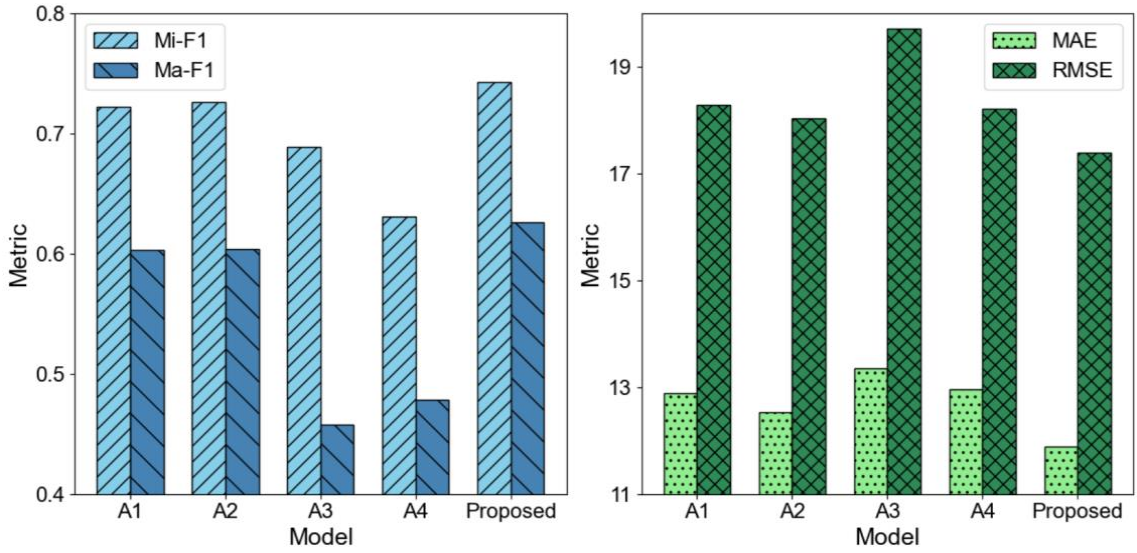

Figure 4. Results of ablation experiments on the Porto dataset for MSRFormer

The results of the ablation experiments indicate that removing the multi-scale spatial interaction extraction module (A3: Ours-w/o-MSIF) results in the most significant decline in performance, particularly in the traffic speed inference task. This finding highlights the importance of this module, as it effectively captures spatial features at various scales and is crucial to the model's success. The contribution of the residual-based multiscale fusion module (A4: Ours-w/o-RMD) is also noteworthy; its removal leads to a considerable drop in model performance, underscoring its role in maintaining stability and facilitating deep feature capture. The spatial flow convolution module (A1: Ours-w/o-SFC) proves to be essential for managing complex spatial dependencies and nonlinear features. Although the interaction feature-based region division module (A2: Ours-w/o-IRP) is important, its impact is not as pronounced as that of the multi-scale feature extraction and residual fusion modules in certain tasks.

*5.4 Scale Selection of Spatial Interactions*

The order of hops in road transfers is a key parameter for extracting multi-scale spatial interaction features, as it directly influences how the road network is divided at different scales. Our experiments established the relationship between road transfer orders and spatial interaction scales for road networks in Porto and San Francisco. To validate our findings, we conducted several experiments to observe how the selection of orders affects model performance by varying the range of orders associated with different scales. For the Porto dataset, we empirically assign $k_S$ to 1 as the representative order for small-scale spatial interaction, since there is no significant difference in the regional divisions when assigning small scale's order $k_S$ to 1 and 2. Therefore, setting $k_S=1$ allows the model to achieve better discrimination between small and medium scale settings, thereby validating the effectiveness of each scales' parameter choices. As for the San Francisco dataset, we used both $k_S=1$ and $k_S=2$ for small scale for sensitivity analysis.

Table 4. Sensitivity analysis of the order $k$ for MSRFormer's performances on road label classification (**RLC**) and traffic inference (**TI**).

(a) Model Performance on the Porto Dataset

| RLC↑, TI↓ / $k_S$, $k_M$ | | $k_L$ 5 | | 6 | | 7 | | 8 | | 9 | |
|---|---|---|---|---|---|---|---|---|---|---|---|
| 1 | 9 | 0.710 | 17.600 | / | / | / | / | / | / | / | / |
| | 5 | / | / | 0.739 | 17.448 | 0.742 | 17.434 | 0.744 | 17.409 | **0.748** | **17.394** |
| | 4 | / | / | 0.737 | 17.466 | 0.738 | 17.451 | 0.741 | 17.427 | 0.742 | 17.420 |
| | 3 | 0.733 | 17.504 | 0.735 | 17.484 | 0.737 | 17.463 | 0.739 | 17.446 | 0.740 | 17.438 |
| | | (3th scale) | | / | / | (4th scale) | | / | / | 0.745 | 17.430 |
| 5 | 7 | / | / | / | / | / | / | / | / | 0.730 | 17.439 |

(b) Model Performance on the San Francisco Dataset

| RLC↑, TI↓ / $k_S, k_M$ | | $k_L$ = 1 | | 5 | | 7 | | 8 | | 9 | |
|---|---|---|---|---|---|---|---|---|---|---|---|
| 1 | 3 | / | / | (3th scale) | | (4th scale) | | / | / | 0.802 | 10.623 |
| | 4 | / | / | / | / | **0.803** | **10.513** | 0.796 | 10.629 | / | / |
| | 5 | / | / | / | / | 0.797 | 10.695 | 0.794 | 10.752 | / | / |
| | 6 | / | / | / | / | 0.789 | 10.680 | 0.784 | 10.740 | / | / |
| 2 | 4 | / | / | / | / | <u>0.801</u> | <u>10.520</u> | 0.798 | 10.635 | / | / |
| | 5 | / | / | / | / | 0.800 | 10.700 | 0.788 | 10.758 | / | / |
| | 6 | / | / | / | / | 0.786 | 10.690 | 0.790 | 10.750 | / | / |
| 7 | 4 | 0.773 | 11.023 | / | / | / | / | / | / | / | / |

Note: The green, pink cells indicate using order choices conforms or violate scale-order relationships respectively. The cells with '/' indicate not attempted order choices, while cells with darker color indicates having better performance. The best and 2nd best results are marked in **bold** and <u>underline</u>.

The experimental results illustrated in Table 4 show that on the Porto dataset, the three scales of order selection ($k_S$=1, $k_M$=5, $k_L$=9) demonstrate optimal performance in road type recognition and traffic inference tasks. For the small-scale order selection, $k_S$=1 effectively meets the model's requirements for extracting spatial interaction features in local areas. For medium-scale order selection, a range of 3 ≤ $k_M$ ≤ 5 successfully captures medium-distance traffic flow characteristics between urban arterials and surrounding regions, leading to improved model performance. In the case of large-scale order selection, the range of 6 ≤ $k_L$ ≤ 9 effectively captures global features, enhancing the model's ability to represent the macroscopic structure of the road network. When the order selection for multi-scale spatial interaction does not adhere to the correct scale-order correspondence—such as repeating the same scale orders ($k_S$=1, $k_M$=3, $k_L$=5) or ($k_S$=5, $k_M$=7, $k_L$=9), or selecting mismatched combinations like ($k_S$=1, $k_M$=9, $k_L$=5)—the model experiences a significant decline in performance. This illustrates that such combinations fail to accurately capture the structural characteristics of the road network, resulting in degraded performance. On the San Francisco dataset, the optimal order selection was found to be ($k_S$=1, $k_M$=4, $k_L$=7), which yielded the best performance in road type identification and traffic inference tasks. Conversely, model performance was poorest when the scale and order were incorrectly aligned, such as in the combination ($k_S$=7, $k_M$=4, $k_L$=1). This reinforces the importance of selecting appropriate orders within the designated scale ranges to achieve better model performance, highlighting the necessity of determining the correct order for spatial interaction scales.

## 6 Discussions

### 6.1 Feature Fusion Strategy's Impact on the Model Performance

To understand the role of multi-scale spatial interaction feature fusion, we selected three sample regions in Porto and San Francisco to evaluate the impact of different scale feature fusion on road network analysis. These regions, labeled (1), (2), and (3). The region (1) focuses on local neighborhood interaction characteristics, looking specifically at intra-community or intra-block dynamics. The region (2) encompasses broader regional connections, reflecting medium-range interactions between urban arterials. Finally, the region (3) addresses cross-regional connections over longer distances, capturing global long-range interaction characteristics. In our experiments, we considered "small," "small+medium," and "small+medium+large" scales to integrate the spatial interaction features of these sample regions effectively.

As illustrated in Figure 5, the model's performance in region (1) in Porto and San Francisco is limited with the integration of medium- and large-scale spatial interaction features. This limitation arises because the roads within these regions are primarily concentrated in communities or neighborhoods, exhibiting distinct localized spatial interaction characteristics and relatively fixed spatial scales. Consequently, these regions are less influenced by larger-scale spatial interactions. In other words, the spatial extent of small-scale areas remains unchanged even as the spatial interaction range expands. For instance, Figure 6 demonstrates that the boundary range and spatial interaction relationships of subareas (3) and (4) in the Porto neighborhood remain stable across different scales ($k$=1, $k$=5, $k$=9). This implies that the model's performance in such regions relies predominantly on local spatial interactions, rather than on multi-scale features from larger areas. Additionally, the model's performance in medium-scale regions is more complex.

The learning of multi-scale spatial interaction features allows for an expanded spatial interaction range in the region (2), significantly improving the model's task performance, particularly related to medium-scale interaction features. This improvement arises because medium-scale spatial interaction features encompass more roads, connecting arterial and secondary road segments, which enhances the connectivity between regions. As a result, the model's ability to represent regional traffic flow characteristics is strengthened. Furthermore, the region (3) exhibits significantly different characteristics compared to regions (1) and (2). Specifically, region (1) shows a more pronounced performance enhancement in multi-scale learning. This can be attributed to their capacity to improve the representation of global interaction patterns through long-distance connectivity, thereby optimizing overall model performance.

In summary, while multi-scale spatial feature fusion has a more limited impact on small-scale spatial interaction regions, medium- and large-scale regions demonstrate significant advantages in capturing long-distance interaction features as the spatial interaction range increases, especially in large-scale areas. This highlights the importance and effectiveness of multi-scale spatial interaction features across different regions.

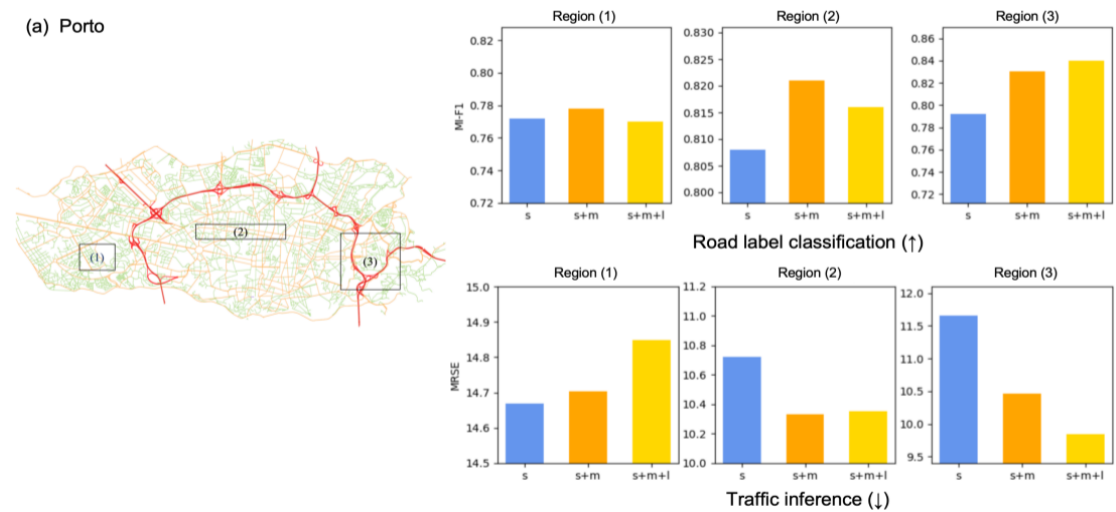

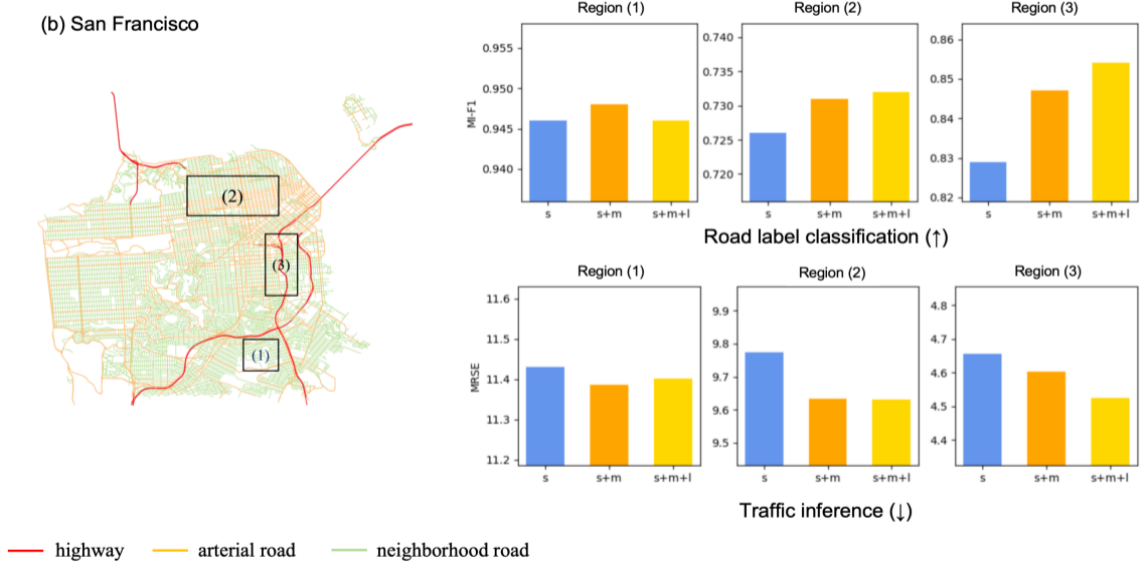

Figure 5. MSRFormer's performances on the two tasks in the sample regions when applying different fusion strategies. s: small-scale feature only, s+m: fusion of small- and medium-scale features, s+m+l: fusion of small-, medium-scale and large-scale features.

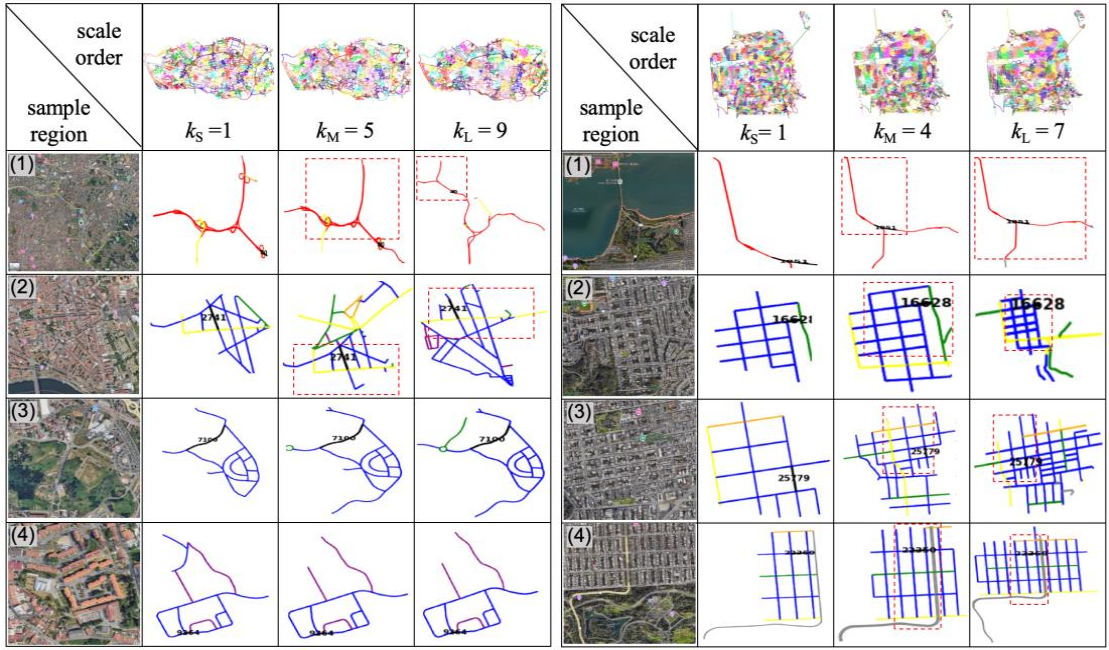

Figure 6. Division of spatial interaction regions of the road network at different scale orders (left: Porto, right: San Francisco). The black digits indicates the same road in different region division while the red-dash-line box indicate the overlap position of the regions with smaller *k*-value.

*6.2 Scale Order's Impacts on the Division of Spatial Interaction Regions*

To understand the differences in the scale effects of spatial interaction features across various experimental areas, we examined the changes in spatial interaction regions at different scales (small, medium, and large) for four selected roads in two experimental areas.

As shown in Figure 7, Porto's road network structure is more decentralized at the small scale ($k_S = 1$), where interactions between roads are confined to localized areas. As the interaction scale increases to medium and high scales ($k_M = 5$ and $k_L = 9$), smaller areas in Porto, such as samples (1) and (2), gradually incorporate road nodes from neighboring areas. Notably, the influence of major roads and highways becomes substantially stronger at these middle and high scales. In contrast, San Francisco displays a more regular and denser division of regions at small scales ($k_S = 1$), characterized by a grid-like road network that facilitates more consistent and dense local interactions among roads. As the scale increases ($k_M = 4$ and $k_L = 7$), the areas of regional interactions expand gradually but with a lesser magnitude, particularly for long-distance interactions, which continue to depend on frequent interactions between residential areas and main roads. This indicates that San Francisco already exhibits strong regional interactions at both low and medium scales, while there is limited enhancement of long-distance interactions at high scales.

In summary, the spatial interactions of different urban road networks are influenced by the hierarchical structure of those networks. The scale effects of spatial interactions also vary: San Francisco's regular road network results in compact regional divisions at low and medium scales, while the high-scale regional divisions show limited expression of long-distance interactions. Conversely, Porto's relatively irregular road network leads to less variation in spatial interaction areas at low and medium scales, while high-scale divisions are similarly limited to the expression of long-distance interactions. Consequently, interaction regions are less variable at low and medium scales, but at high scales, the interaction among arterial roads and highways significantly strengthens, reflecting its hierarchical road network structure. This difference in scale effects results in distinct choices for appropriate scale orders in the two experimental areas.

*6.3 Embedding Analysis*

To investigate the impact of multi-scale spatial interaction features on the construction of embedded representations of road networks, we analyze the road embedding vectors generated by GCN, TrajRNE, and MSRFormer. We compare the representational performance of the embedding vectors from these three models, focusing on the spatial structure of the embedding vectors (see Figure 7(a)) and their similarity (see Figure 7(b)). For this analysis, we selected regions defined by small, medium, and large-scale spatial interaction matrices, referred to as regions S, M, and L, respectively, within the Porto road network (see Figure 7(a)). We embedded the Porto road network using each of the three models and visualized the results using the dimension reduction technique t-SNE. As shown in Figure 7(a), the embedding distributions of the three sample regions scattered across the embedding space of GCN and TrajRNE. In contrast, the embedding distribution from MSRFormer compactly clusters roads with similar spatial interaction scales. This suggests that MSRFormer is more effective in distinguishing between different spatial interaction scales of roads.

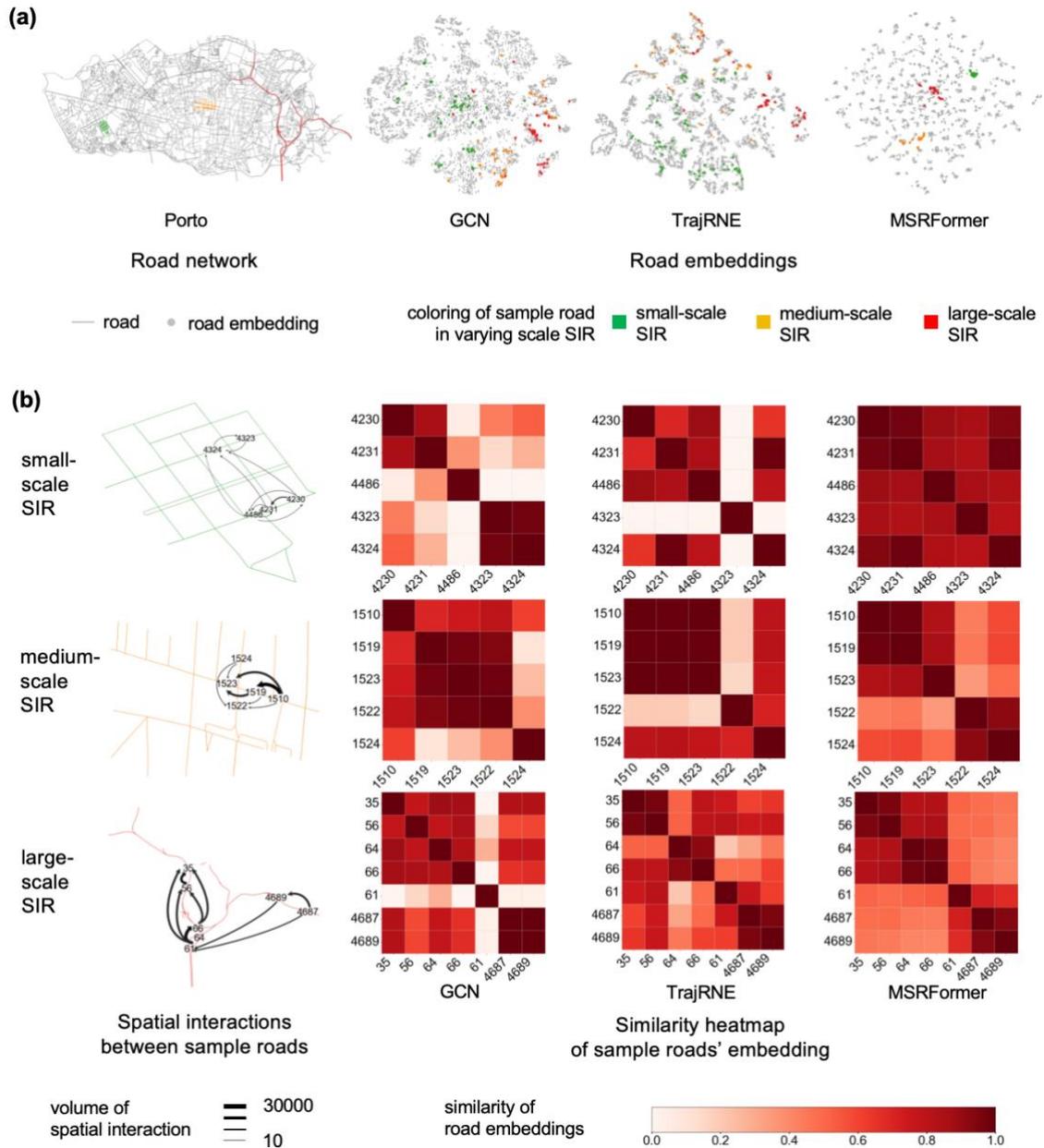

Figure 7. Embedding analysis of GCN, TrajRNE and MSRFormer on the Porto dataset. (a) Distribution patterns of embedded roads from varying scales of spatial interaction regions (SIR), (b) Similarity of learned road embeddings compared with roads' structure and spatial interactions.

To further analyze the embedding characteristics of road network nodes within the same spatial interaction scale, we selected embedding vectors from several roads in the sample regions for a similarity analysis (as depicted in Figure 7(b)). In the small-scale spatial interaction region (SIR), road segment interactions exhibit short distances and homogeneity (i.e., spatial interactions have identical volumes). The MSRFormer model effectively captures these local spatial interactions, resulting in distinguishing similarity patterns (i.e., cells with darker color) with small similarity differences. In the medium-scale SIR, road segment interaction patterns show significant heterogeneity with roads with distinguished volume of spatial interactions. The MSRFormer model effectively differentiates between these segments with distinct interaction patterns, creating two separate clusters on the node embedding similarity heat map (shown in Figure 7(b) with MSRFormer nodes 1510, 1519, 1523 and nodes 1522, 1524). In the large-scale SIR, even though road segments (node 61 and nodes 4687, 4689) are far apart from each other, they remain highly clustered in the representation space when using the MSRFormer model. This demonstrates the model's strength in capturing long-range spatial interactions, as the node embedding vectors learned via MSRFormer for these long distances still show similar characteristics.

In contrast, GCN and TrajRNE overly rely on road network neighborhood topology and low-order spatial interaction aggregation, which limits their ability to effectively learn interaction patterns in heterogeneous spatial interaction regions.

**7 Conclusions**

This paper introduces a representation learning framework for road networks, called MSRFormer. This framework utilizes multi-scale spatial interaction features to improve the model's capacity to represent complex urban road networks. Comprehensive experiments conducted on two urban road network and trajectory datasets show that MSRFormer effectively addresses issues such as noise introduction and feature smoothing in GNNs when using wider and deeper feature aggregators. It also proves effective in modeling flow heterogeneity and long-distance spatial dependencies within the road network. Moreover, it exhibits strong generalization capabilities across various data scenarios. An ablation study further underscores the importance of the multi-scale spatial interaction feature extraction and fusion module in improving model performance.

This paper presents a practical framework for building task-agnostic road network representation models and offers an insightful investigation into the scale effects of geographic embedding models that handle spatial interactions. Future research could further explore the multi-scale effects of spatial interactions and thoroughly examine the complexities of these interactions within urban road networks. This includes focusing on the semantics of interaction behaviors, such as travel patterns between different functional urban regions, as well as considering multi-modal transportation methods, like urban transit involving both roads and railways. Moreover, more advanced contrastive learning techniques could be used to further improve the performance of MSRFormer. Such studies would provide valuable theoretical insights and practical guidance for the development of urban applications, including intelligent transportation systems and smart city initiatives.

**Acknowledgments**

The authors would like to thank Mr. Stefan Schestakov for his helpful discussion on the implementation of TrajRNE and sharing the data and codes, and Zhangxiang Lin for helping with the supplementary experiments during the review process. The authors would also like to thank the anonymous reviewers for their valuable comment that improved the manuscript.

**Disclosure statement**

The author reports no potential conflicts of interest.

**Funding**

This work was funded by National Natural Science Foundation of China under Grant No.42130112, No.42371479, No.41901335; and China's National Key R&D Program No.2017YFB0503500.

**Data availability statement**

The data that support the findings of this study are openly available online: 1) Porto. https://www.kaggle.com/competitions/pkdd-15-taxi-trip-time-prediction-ii; 2) San Francisco. https://ieee-dataport.org/open-access/crawdad-epflmobility.

**CRediT authorship contribution statement**

Jian Yang: conceptualization, methodology, writing – original draft, writing – review & editing, visualization, software, supervision, funding acquisition. Jiahui Wu: methodology, writing – original draft, writing – review & editing, software. Li Fang: writing – review & editing. Hongchao Fan: writing – review & editing. Bianying Zhang: data curation, validation. Huijie Zhao: data curation, validation. Guangyi Yang: data curation, validation. Xiong You: writing – review & editing, funding acquisition.